\documentclass[journal]{IEEEtran}
\usepackage{amsmath,amsfonts}
\usepackage{algorithm}
\usepackage{array}
\usepackage[caption=false,font=normalsize,labelfont=sf,textfont=sf]{subfig}
\usepackage{textcomp}
\usepackage{stfloats}
\usepackage{url}
\usepackage{verbatim}
\usepackage{graphicx}
\usepackage{cite}
\usepackage{amsmath,amsfonts,bm}

\def\eqref#1{equation~\ref{#1}}

\def\1{\bm{1}}

\def\vw{{\bm{w}}}

\DeclareMathAlphabet{\mathsfit}{\encodingdefault}{\sfdefault}{m}{sl}
\SetMathAlphabet{\mathsfit}{bold}{\encodingdefault}{\sfdefault}{bx}{n}

\def\gA{{\mathcal{A}}}

\def\gG{{\mathcal{G}}}

\def\gS{{\mathcal{S}}}

\usepackage{url}
\usepackage{graphicx}
\usepackage{silence}
\usepackage{subcaption}
\usepackage{algpseudocode} 
\usepackage{amsmath} 
\usepackage{float}
\usepackage{array,booktabs,makecell,multirow}
\usepackage{xspace}
\usepackage{xcolor}

\usepackage{thmtools, thm-restate}
\newtheorem{theorem}{Theorem}[section]
\newtheorem{proposition}[theorem]{Proposition}

\newtheorem{definition}[theorem]{Definition}

\newcommand{\algo}{HOLA\xspace} 

\newcommand{\algoR}{$\text{HOLA}_R$\xspace}

\begin{document}

\title{Multi-Robot Open Adaptive Teaming Across \\ Unseen Environments, Partners, and Scales}

\author{
\IEEEauthorblockN{
Yang Li\textsuperscript{1,*},
Feng Xue\textsuperscript{2},
Fan Mo\textsuperscript{3},
Yunhao Liu\textsuperscript{4},
Jianhong Wang\textsuperscript{5},\\
Ying Wen\textsuperscript{1}, 
Qingrui Zhang\textsuperscript{2},
Shaoshuai Mou\textsuperscript{6},
Wei Pan\textsuperscript{7,*}
}

\IEEEauthorblockA{
\textsuperscript{1}Shanghai Jiao Tong University \quad
\textsuperscript{2}Sun Yat-sen University \quad
\textsuperscript{3}National University of Singapore \\
\textsuperscript{4}Genisom AI \quad
\textsuperscript{5}University of Bristol \quad
\textsuperscript{6}Purdue University \quad
\textsuperscript{7}Newcastle University \\
\textsuperscript{*}Corresponding author: \texttt{wei.pan2@newcastle.ac.uk, yang.li.cs@sjtu.edu.cn}
}
}


\markboth{Journal of \LaTeX\ Class Files,~Vol.~14, No.~8, August~2021}%
{Shell \MakeLowercase{\textit{et al.}}: A Sample Article Using IEEEtran.cls for IEEE Journals}


\maketitle

\begin{abstract}
Deploying robot teams in the real world requires simultaneous adaptation to unseen environments, unknown partners, and varying team sizes, yet existing approaches often address these challenges in isolation under the closed-world assumption of fixed teammates. We formalize this as open adaptive multi-robot teaming and propose a hypergraphic-form game formulation that captures team-level cooperative relationships beyond pairwise interactions, providing a principled foundation for coordination structure inference when team composition changes dynamically within episodes. Unlike graph neural network architectures, this is a game-theoretic construct for modeling strategic interactions and payoff structures among agents. Building on this formulation, we develop the Hypergraphic Open-ended Learning Algorithm (\algo), which progressively expands partner and environment diversity during training rather than optimizing for fixed configurations. Evaluated on cooperative pursuit with multi-drone and multi-quadruped platforms, \algo outperforms all baselines across all three adaptability dimensions. Learned policies transfer directly to physical hardware without fine-tuning, with successful deployments on Crazyflie and Zsibot L1 platforms confirming robust real-world coordination in novel environments with unseen teammates.
\end{abstract}


\section{Introduction}

Multi-robot systems are rapidly transitioning from controlled laboratory settings to unpredictable real-world environments such as disaster zones, dynamic warehouses, and surveillance networks, where the fundamental assumptions underlying conventional coordination mechanisms break down~\cite{chung2011search,ZhangDACOOP2023,queralta2020collaborative}.
In these settings, robots cannot rely on familiar terrain, known teammates, or stable team configurations. 
A search-and-rescue drone may lose communication with its squad and must instantly coordinate with an unfamiliar ground unit; a warehouse robot must seamlessly integrate a newly deployed agent mid-task; a border patrol team must reorganize on-the-fly as units fail or reinforcements arrive. 
These scenarios share a common challenge that existing approaches fail to address: robots must simultaneously generalize across {unseen environments}, coordinate with {unknown partners}, and scale to {varying team sizes}—all in real time, without retraining. We formalise this challenge as the \textbf{open adaptive multi-robot teaming} problem.

\begin{figure}[ht]
    \centering
    \includegraphics[width=\linewidth]{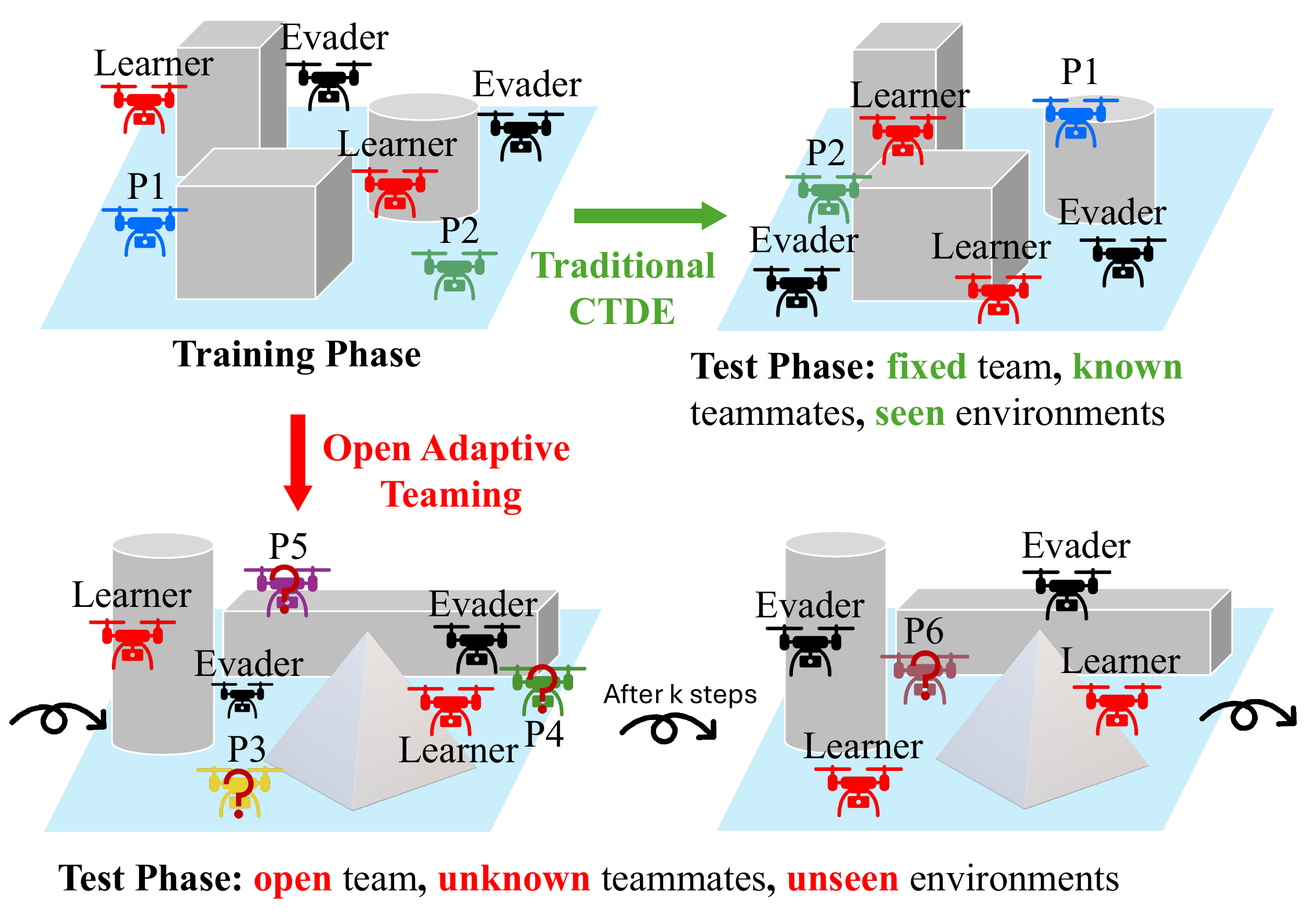}
    \caption{
Comparison of Traditional CTDE and Open Adaptive Teaming Paradigm. In conventional CTDE (top right), the test phase maintains a fixed team composition with known teammates in previously seen environments. In contrast, open adaptive teaming (bottom) presents significantly greater challenges: agents must coordinate in open teams with unseen teammates and dynamic scale in unseen environments during online interaction.
}
    \label{fig:intro}
\end{figure}

Fig.~\ref{fig:intro} contrasts the dominant learning paradigm with the open adaptive teaming challenge. 
Centralized training with decentralized execution (CTDE)~\cite{lowe2017multi, rashid2018qmix, yu2022surprising} trains a fixed set of agents jointly, allowing them to develop shared \emph{coordination conventions}—implicit protocols for cooperation that emerge through co-training~\cite{hu2020other}. 
At deployment, CTDE assumes that the same agents operate together in environments similar to those seen during training (Fig.~\ref{fig:intro}, top). 
This closed-world assumption has enabled impressive results in tasks from collision avoidance~\cite{long2018towards} to formation control~\cite{li2020graph}, but fundamentally limits real-world applicability.
Open adaptive teaming breaks all three assumptions simultaneously 
(Fig.~\ref{fig:intro}, bottom): the environment changes 
(novel layouts and obstacles), the partners are unknown (agents 
$P_3$--$P_6$ were never encountered during training), and team 
composition evolves dynamically {within} episodes rather 
than only across them---agents join or depart after $k$ steps, 
requiring continuous re-coordination during task execution.
In this setting, coordination conventions learned through co-training become liabilities rather than assets—robots must infer effective cooperation strategies on-the-fly with partners whose behaviors they have never observed.

Multi-agent learning in machine learning field has made significant progress on coordination challenges in simulation, including zero-shot coordination and ad hoc teamwork  with unseen partners~\cite{hu2020other,li2023cole,stone2010ad,wang2024ciao}. 
However, these advances rely on assumptions that do not hold in physical multi-robot systems: discrete action spaces, deterministic transitions, and stable communication.
When deployed on real robots, where agents must contend with continuous dynamics, sensor noise, partial observability, and the possibility of communication loss or hardware failure, existing methods exhibit brittle performance.
Domain randomization~\cite{tobin2017domain} has enabled robust sim-to-real transfer for single-robot control, yet extending this paradigm to multi-robot coordination introduces a fundamentally harder problem: the environment is no longer the only source of uncertainty, as teammate behaviors themselves become non-stationary and unpredictable.
This interplay between environmental and social uncertainty creates theoretical challenges that cannot be addressed by straightforward extensions of existing frameworks, demanding new formulations tailored to the multi-robot setting.

To address this challenge, we propose the \textbf{H}ypergraphic \textbf{O}pen-ended \textbf{L}earning \textbf{A}lgorithm (\textbf{HOLA}), a unified framework for open adaptive multi-robot teaming.
The core of our approach is a hypergraphic-form game formulation that captures dynamic, team-level cooperative relationships among agents. Unlike pairwise interaction models that decompose coordination into agent-agent edges, hypergraphs naturally represent higher-order dependencies where three or more robots must act jointly to achieve a subtask, such as cooperative manipulation, encirclement, or distributed sensing. Importantly, this formulation is a game-theoretic construct for modeling strategic interactions and payoff structures among agents, not a neural architecture based on graph neural networks. This representation enables principled reasoning about how individual agents contribute to team performance across varying team compositions, providing a theoretically grounded mechanism for credit assignment in open teams. Building on this formulation, we develop an open-ended learning framework that progressively expands the diversity of training partners and environments, rather than optimizing for fixed team configurations.

We evaluate HOLA on multi-robot pursuit, a canonical coordination task that requires real-time tracking and interception of dynamic evaders~\cite{chung2011search}.
Pursuit serves as an ideal testbed for open adaptive teaming: effective capture strategies demand continuous adaptation to environmental geometry, teammate capabilities, and team configuration, making it a rigorous benchmark for all three dimensions of adaptability.
Our evaluation protocol systematically tests generalization along each dimension.
For environmental generalization, we train on a fixed set of layouts and evaluate on unseen obstacle configurations.
For partner adaptability, agents must coordinate zero-shot with teammates trained using different algorithms and reward structures.
For team scalability, we vary the number of pursuers within episodes, requiring agents to maintain effective coordination as teammates join or depart.
We compare HOLA against multi-agent reinforcement learning baselines (self-play, MAPPO~\cite{mappo}) and population-based training methods designed for zero-shot coordination (PBT~\cite{jaderberg2017population}, FCP~\cite{strouse2021collaborating}).
Results demonstrate that HOLA consistently outperforms all baselines across the three generalization dimensions.
Beyond simulation, we deploy learned policies directly onto physical platforms, including multi-drone systems and quadruped robots, without policy fine-tuning.
Successful real-world coordination with unseen teammates in novel environments validates the sim-to-real transferability of our approach.

Our contributions are threefold:
\begin{itemize}
   \item We formalize the open adaptive multi-robot teaming problem, which requires simultaneous generalization across unseen environments, unknown partners, and varying team sizes. We introduce a hypergraphic-form game representation that captures dynamic team-level coordination structures, providing a theoretical foundation for learning in open teams.
    \item We propose HOLA, a hypergraphic open-ended learning algorithm that integrates the hypergraphic game formulation with an adaptive training framework. HOLA progressively expands partner and environment diversity during training, enabling agents to develop robust coordination strategies without optimizing for fixed team configurations.
    \item  We validate HOLA through extensive simulations and real-world deployments on multi-drone and multi-quadruped platforms. Results confirm that learned policies generalize across all three dimensions simultaneously, achieving robust coordination in novel environments, with unseen teammates, and under dynamic team compositions.
\end{itemize}

The remainder of this paper is organized as follows. Section~\ref{app:related} surveys related work on multi-robot pursuit and open adaptive teaming. Section~\ref{sec:preli} introduces the mathematical preliminaries and problem formulation. Section~\ref{sec:HOLA} presents the proposed hypergraphic open-ended learning algorithm and its theoretical foundations. Section~\ref{sec:exp_setting} describes the experimental setup, and Section~\ref{sec:exp_res} reports the results. Section~\ref{sec:conclusion} concludes the paper with a discussion of limitations and future directions.

\section{Related Work}
\label{app:related}
\paragraph{Multi-robot pursuit}
Multi-robot pursuit represents a fundamental challenge in cooperative robotics, where multiple agents must coordinate to capture or intercept one or more evaders. Existing approaches can be categorized into three main paradigms: rule-based heuristic methods, differential game theoretical methods, and learning-based methods.
Rule-based heuristic methods are primarily inspired by biological hunting behaviors observed in nature. These approaches typically employ artificial attractive and repulsive forces to coordinate pursuit actions~\cite{madden_multi-robot_2010, muro_wolf-pack_2011, angelani_collective_2012}. For instance, Janosov et al.\cite{janosov_group_2017} designed predictive attraction forces, obstacle repulsive forces, and teammate repulsive forces to achieve encirclement. Similarly, Pierson et al.\cite{pierson_intercepting_2017} proposed a Voronoi tessellation-based method suitable for ground vehicle coordination. While computationally efficient, these manually designed rules are limited by the designer's observations and experiences, restricting their applicability to complex, dynamic scenarios.
Differential game theoretical methods formulate pursuit as an optimisation problem, deriving theoretically optimal strategies through game-theoretic analysis~\cite{mu_survey_2023, garcia_geometric_2017, kothari_cooperative_2017, hayoun_two--one_2017}. Hayoun et al.~\cite{hayoun_two--one_2017} modelled two-on-one pursuit as a zero-sum game and obtained analytical solutions. Kothari et al.~\cite{kothari_cooperative_2017} considered pursuers and evaders as nonholonomic systems and employed model predictive control to minimise the evader's safe zone. These methods maximise utility functions based on the Hamilton-Jacobi-Bellman equation but require precise state transition models. Consequently, their performance degrades significantly under environmental uncertainty and model inaccuracies.
Learning-based methods have emerged to address the limitations of rule-based and game-theoretic approaches by learning cooperative pursuit strategies from data~\cite{matignon_hysteretic_2007, li_robust_2019, qi_cascaded_2024, de_souza_decentralized_2021}. De Souza et al.\cite{de_souza_decentralized_2021} employed curriculum learning and parameter sharing to train agents that approach, decelerate, and surround evaders—exhibiting emergent intelligent behaviors similar to biological hunting. Zhang et al.\cite{zhang_multi-agent_2022} introduced attention mechanisms to enhance agent-environment interaction, demonstrating scalability in 100-on-100 pursuit scenarios.

Recent work has begun exploring generalization capabilities in pursuit-evasion problems. Zhang et al.\cite{ZhangDACOOP2023} integrated rule-based strategies into reinforcement learning to improve data efficiency and generalization across environments. DualCL\cite{chen2023taskflex} enhanced capture strategies through progressive parameter adjustment and diverse scenario generation for zero-shot transfer to unseen environments. Chen et al.~\cite{chen2024multi} developed a deep reinforcement learning approach enabling UAVs to perform online planning in unknown environments under partial observability and dynamic constraints. However, these approaches primarily address environmental adaptability while assuming fixed team compositions with known teammate policies. In contrast, our work tackles the broader challenge of open adaptive teaming, enabling robots to coordinate effectively with unknown partners across varying team scales without prior knowledge of teammates or task scenarios.

\paragraph{Open adaptive teaming}
Three research paradigms address coordination with previously unseen teammates: ad hoc teamwork, zero-shot coordination, and dynamic team composition. Stone et al.~\cite{stone2010ad} formalized ad hoc teamwork as collaboration without pre-coordination, with subsequent work developing algorithms for teammate modeling~\cite{barrett2017making}, graph-based architectures for dynamic team composition~\cite{rahman2023general}, and cooperative game-theoretic frameworks~\cite{wang2024open, wang2025shapley}. Unlike ad hoc teamwork which allows for online adaptation, ZSC requires agents to coordinate effectively at first contact without any prior interaction. Foundational approaches include Other-Play~\cite{hu2020other}, which exploits environmental symmetries, and Off-Belief Learning~\cite{hu2021off}, which trains grounded policies invariant to training randomness. Population-based methods~\cite{strouse2021collaborating, zhao2023maximum, li2023cooperative, li2024tackling} have proven particularly effective, training against diverse partners or checkpoints to improve generalization. Dynamic team composition research focuses on handling variable team sizes through entity-based representations~\cite{hu2021updet}, transformer architectures~\cite{wen2022multi}, and curriculum learning~\cite{long2020evolutionary, wang2023skilled}.

However, while these paradigms have advanced coordination theory, they predominantly address individual dimensions of adaptability in isolation, leaving the challenge of simultaneous multi-dimensional adaptation unresolved. More critically, existing approaches lack a unified framework integrating partner adaptation, environmental generalization, and team scalability. Furthermore, most advances in open adaptive teaming remain confined to video game benchmarks (Hanabi, Overcooked, StarCraft) with limited real-world robotic validation~\cite{chen2024multi, ZhangDACOOP2023}, creating a substantial gap between coordination theory and practical deployment.
In contrast, our work establishes a principled framework for open adaptive teaming in multi-robot systems that unifies all three dimensions of adaptability within a single coherent approach. 

\section{Preliminaries and Problem Formulation}
\label{sec:preli}
We first establish the mathematical preliminaries underlying our approach, then formally define the multi-robot open adaptive teaming problem with its key challenges and requirements.

\subsection{Preliminaries}
\paragraph{Normal-form Game.} A normal-form game $\mathcal{G} = (N, \{S_i\}_{i=1}^{|N|}, \{u_i\}_{i=1}^{|N|})$ comprises a finite set of players $N$, indexed by $i \in \{1, 2, \ldots, |N|\}$; a set of strategies $S_i$ for each player $i \in N$; and reward functions $u_i: \prod_{j \in N} S_j \rightarrow \mathbb{R}$.
Moreover, $\sigma = (\sigma_1, \ldots, \sigma_n)$ denotes the strategy profile, assigning a strategy to each player. Here, $\sigma_i \in \Delta(S_i)$ represents a mixed strategy for player $i$, where $\Delta(S_i)$ denotes the probability distribution over the set $S_i$.
The subscript $-i$ is used to denote all players excluding player $i$. For instance, $\sigma_{-i}$ denotes a strategy profile for all players except player $i$.
We say a best response (BR) of player $i$ to strategy $\sigma_{-i}$ if $BR(\sigma_{-i})=\arg \max_{\sigma_i \in \triangle(S_i)}$. 

\paragraph{Hypergraph} 
A hypergraph, denoted as $\mathcal{G} = (\mathcal{V}, \mathcal{E}, \mathbf{W})$, captures higher-order relationships among multiple entities simultaneously, a capability absent in conventional graphs that are restricted to pairwise connections. Here, $\mathcal{V}$ and $\mathcal{E}$ denote the sets of vertices and hyperedges, respectively. Each hyperedge $e \in \mathcal{E}$ connects an arbitrary subset of vertices and carries a weight $w(e)$ indicating the strength of that relationship. The weight information is consolidated in the diagonal matrix $\mathbf{W}$, where $\mathrm{diag}(\mathbf{W}) = \bigl(w(e_1), w(e_2), \ldots, w(e_{|\mathcal{E}|})\bigr)$.

\paragraph{Partially Observable Markov Decision Process}
We formalize each robot's decision-making as a Partially Observable Markov Decision Process (POMDP), a tuple $\langle \mathcal{S}, \mathcal{A}, \mathcal{T}, \mathcal{R}, \Omega, \mathcal{O}, \gamma \rangle$, where $\mathcal{S}$ is the state space, $\mathcal{A}$ is the action space, $\mathcal{T}: \mathcal{S} \times \mathcal{A} \times \mathcal{S} \rightarrow [0,1]$ is the transition function, $\mathcal{R}: \mathcal{S} \times \mathcal{A} \rightarrow \mathbb{R}$ is the reward function, $\Omega$ is the observation space, $\mathcal{O}: \mathcal{S} \times \mathcal{A} \rightarrow \Omega$ is the observation function, and $\gamma \in [0,1)$ is the discount factor. At timestep $t$, an agent receives partial observation $o_t$, executes action $a_t$ according to policy $\pi: \Omega \rightarrow \mathcal{A}$, receives reward $r_t$, and transitions to state $s_{t+1}$. The objective is to find an optimal policy $\pi^*$ that maximizes expected cumulative return $\mathbb{E}_{\pi}[\sum_{t=0}^{\infty} \gamma^t r_t]$.

For multi-robot coordination, we employ a Decentralized POMDP (Dec-POMDP) with $n$ agents. Each agent $i$ observes $o_t^i \in \Omega^i$, executes action $a_t^i \in \mathcal{A}^i$ via local policy $\pi^i: \Omega^i \rightarrow \mathcal{A}^i$, and the joint action $\mathbf{a}_t = (a_t^1, \ldots, a_t^n)$ yields team reward $\mathcal{R}(s_t, \mathbf{a}_t) = \sum_{i=1}^n \mathcal{R}^i(s_t, \mathbf{a}_t)$.

\subsection{Problem Formulation}
\label{sec:problem_formulation}

\begin{definition}[Open Adaptive Teaming for Multi-Robot Systems]
An Open Adaptive Teaming problem for multi-robot systems is defined by a tuple $\langle \mathcal{L}, \mathcal{U}, \gS, \gA, \Theta, \mathcal{E}, \Phi, R, P, \mathcal{P}_{\text{env}} \rangle$, where $\mathcal{L}$ represents the set of learnable robot agents, $\mathcal{U}$ represents the set of uncontrolled partner agents, $\gS$ is the state space, $\gA$ is the action space, $\Theta$ is the type space characterizing partner behaviors, $\mathcal{E}$ is the environment space, $\Phi$ is the robot constraint set encoding physical and operational limitations, $R$ is the reward function, $P$ is the transition function, and $\mathcal{P}_{\text{env}}$ is the environment distribution. Let $\mathcal{C} = \mathcal{L} \cup \mathcal{U}$ denote the complete agent set and $\mathcal{P}$ denote the power set.

The robot constraint set $\Phi = \{\phi_1, \phi_2, \ldots, \phi_m\}$ encapsulates various physical and operational constraints inherent to robotic platforms, including but not limited to dynamics constraints (velocity and acceleration limits), safety constraints (collision avoidance, workspace boundaries), communication constraints (range and bandwidth limitations), sensing constraints (field-of-view, occlusion), energy constraints (battery limits, power consumption), and task-specific operational constraints.

To handle variable team size, we define a joint agent-action space 
\begin{equation}
    \mathbf{A}_{\mathcal{C}} = \{a | a \in \mathcal{P}(\mathcal{C} \times \gA), \forall (i, a^i), (j, a^j) \in a : i = j \Rightarrow a^i = a^j\},
\end{equation}
where elements $a \in \mathbf{A}_{\mathcal{C}}$ are referred to as \textit{joint agent-actions}. Similarly, we define a joint agent-type space
\begin{equation}
    \mathbf{\Theta}_{\mathcal{C}} = \{\theta | \theta \in \mathcal{P}(\mathcal{C} \times \Theta), \forall (i, \theta^i), (j, \theta^j) \in \theta : i = j \Rightarrow \theta^i = \theta^j\},
\end{equation}
where $\theta \in \mathbf{\Theta}_{\mathcal{C}}$ is the \textit{joint agent-type} and $\theta^i$ denotes the type of agent $i$ in $\theta$.

The reward function is decomposed as $R: \gS \times \mathbf{A}_{\mathcal{C}} \times \mathcal{E} \times \Phi \mapsto \mathbb{R}$, with
\begin{equation}
    R(s, a, e, \phi) = R^{\text{task}}(s, a, e) + R^{\text{constraint}}(s, a, \phi),
    \label{eq:reward_decomposition}
\end{equation}
where $R^{\text{task}}(s, a, e)$ rewards task performance and $R^{\text{constraint}}(s, a, \phi)$ penalizes constraint violations:
\begin{equation}
    R^{\text{constraint}}(s, a, \phi) = -\sum_{i=1}^{m} \lambda_i \mathcal{V}_i(s, a, \phi_i),
    \label{eq:constraint_penalty}
\end{equation}
where $\mathcal{V}_i(s, a, \phi_i)$ quantifies the violation of constraint $\phi_i \in \Phi$ and $\lambda_i > 0$ are penalty weights balancing task performance and constraint satisfaction.

The transition function $P: \gS \times \mathbf{A}_{\mathcal{C}} \times \mathcal{E} \times \Phi \mapsto \Delta(\gS \times \mathbf{\Theta}_{\mathcal{C}})$ determines the probability distribution over next states and joint agent-types, given the current state, joint agent-actions, environment, and robot constraints, where $\Delta(X)$ denotes the set of all probability distributions over $X$. The robot constraints $\Phi$ influence both feasible state transitions and team composition dynamics.

At the start of an episode, an environment $e \sim \mathcal{P}_{\text{env}}(\mathcal{E})$ is sampled, along with an initial state $s_0$ and an initial set of agents $\mathcal{C}_0 \subseteq \mathcal{C}$ with associated types $\theta_0^i$ for $i \in \mathcal{C}_0$, from a starting distribution $P_0 \in \Delta(\gS \times \mathbf{\Theta}_{\mathcal{C}})$. At state $s_t$, agents $\mathcal{C}_t \subseteq \mathcal{C}$ with types $\theta_t^i, i \in \mathcal{C}_t$ exist in the environment and choose their actions $a_t^i$ by sampling from their policies $\pi^i$. The learner receives a reward computed through $R(s_t, a_t, e, \phi)$, where $a_t \in \mathbf{A}_{\mathcal{C}_t}$ denotes the joint agent-actions. The next state $s_{t+1}$ and the next set of existing agents $\mathcal{C}_{t+1}$ with types are sampled from $P$ given $s_t$, $a_t$, $e$, and $\phi$.

The objective is to learn policies $\{\pi^i\}_{i \in \mathcal{L}}$ for learnable robot agents that maximize the expected discounted return while respecting robot constraints:
\begin{equation}
    \mathcal{J} = \mathbb{E}_{e \sim \mathcal{P}_{\text{env}}, \tau \sim P}\left[\sum_{t=0}^T \gamma^t R(s_t, a_t, e, \phi)\right],
    \label{eq:obj_oat}
\end{equation}
where $\tau$ denotes the trajectory and $\gamma \in [0,1)$ is the discount factor. This formulation ensures that learned policies generate robot-feasible behaviors deployable on physical platforms.
\end{definition}
\section{Hypergraphic Open-ended Learning Algorithm}
\label{sec:method}

\begin{figure}[t]
  \centering
    \includegraphics[width=\linewidth]{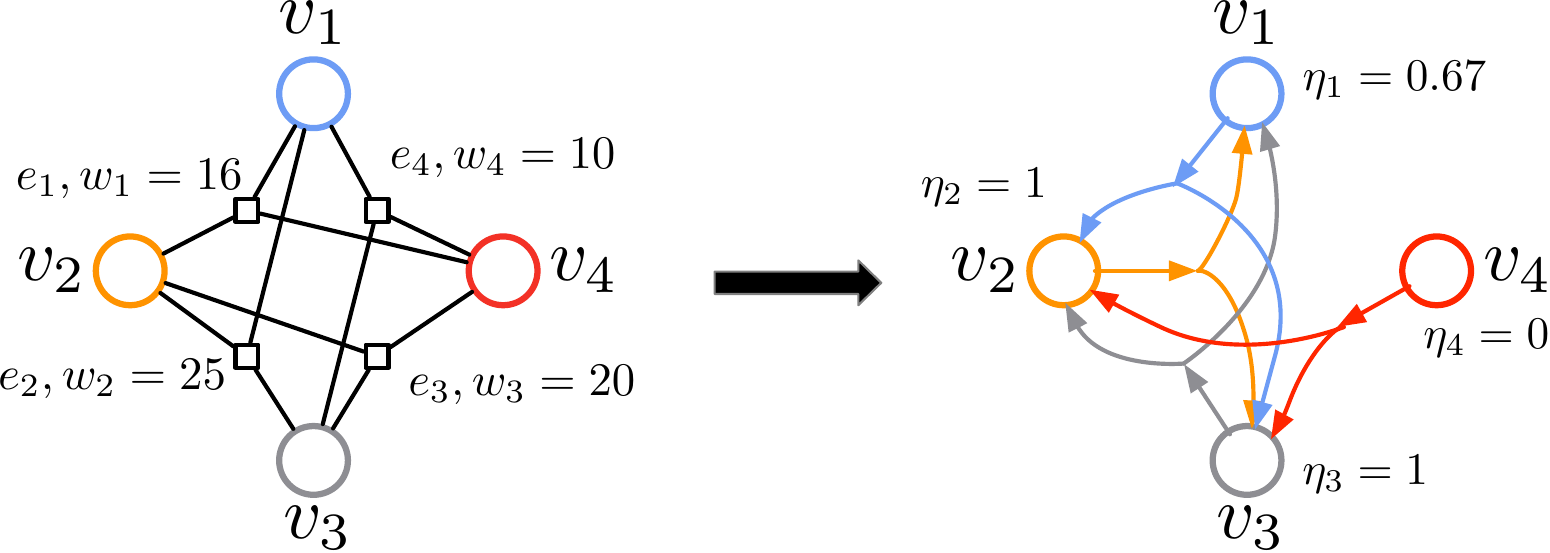} 
    \caption{Schematic illustration of Open Hypergraphic-Form Game (left) and its corresponding preference hypergraph (right). Hyper-preference centrality is computed using in-degree centrality aggregated across hyperedge cardinalities.}
    \label{fig:pref_H}
\end{figure}

\begin{figure*}[t]
    \centering
    \includegraphics[width=\textwidth]{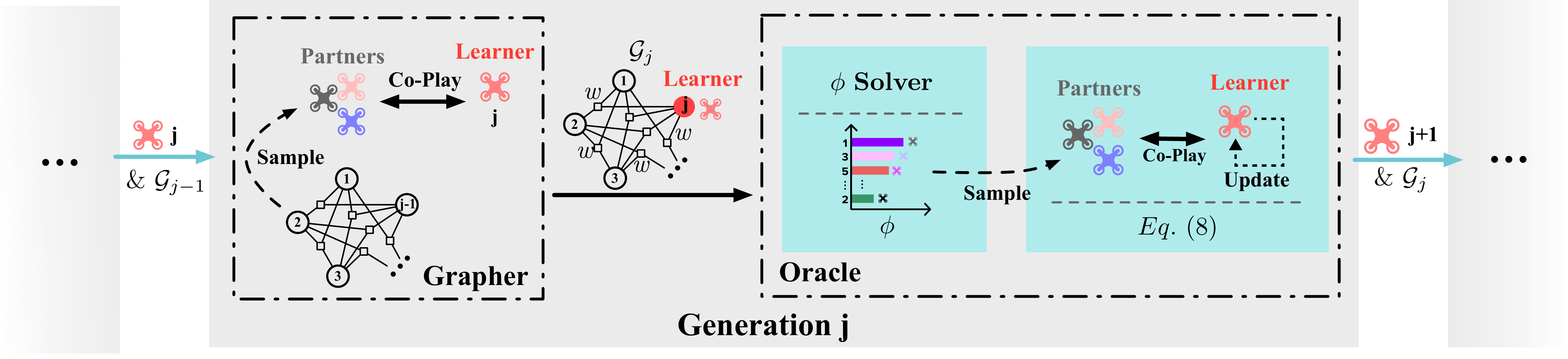}
    \caption{Hypergraphic Open-ended Learning Algorithm: Detailed illustration of a single generation within the open-ended learning phase, including the grapher and oracle modules.}
    \label{fig:model}
\end{figure*}

To address the multi-robot open adaptive learning problem, we introduce a novel approach named the hypergraphic open-ended learning algorithm (\algo). 
In Section~\ref{sec:Pref_H}, we first introduce the preference hypergraph and hyper-preference centrality to model cooperative relationships and assess the coordination ability of each agent within the hypergraph.
In Section~\ref{sec:HOLA}, we provide the details of \algo, as illustrated in Fig.~\ref{fig:model}.

\subsection{Open Preference Hypergraph}
\label{sec:Pref_H}

We propose a hypergraphic-form game formulation to model multi-agent interactions in open adaptive teaming, extending beyond traditional normal-form games that represent pairwise interactions. In multi-robot coordination, the utility of cooperative behaviors often depends on higher-order interactions among multiple agents simultaneously—for instance, successful target encirclement requires coordinated positioning of three or more robots, which cannot be decomposed into pairwise relationships. While normal-form games represent two-player interactions as matrices and have been extended to tensors for multi-player games, they become computationally intractable for variable team sizes and dynamic team compositions. Our hypergraphic-form game leverages hypergraphs—mathematical structures where edges (hyperedges) can connect arbitrary numbers of vertices—to naturally represent these higher-order cooperative relationships. Crucially, this is a \textit{game-theoretic formulation} for modeling strategic interactions and payoff structures, not a neural architecture based on graph neural networks. Each vertex represents an agent's policy (parameterized by a neural network), and each hyperedge captures the joint utility obtained when a specific subset of agents coordinate their actions. This formulation enables efficient reasoning about cooperation patterns across varying team compositions while maintaining tractability as team size changes.

\begin{definition}[Open Hypergraphic-Form Game]
The \textbf{O}pen \textbf{H}ypergraphic-Form Game (OH-Game) $\gG$ is defined by the tuple $(\mathcal{V}, \mathcal{E}, \vw, \mathcal{L})$, where $\mathcal{V}$ is a set of vertices representing agent policies parameterized by neural network weights $\theta^i$; $\mathcal{E} = \bigcup_{l \in \mathcal{L}} \mathcal{E}_l$ is a set of hyperedges with varying cardinalities, where $\mathcal{E}_l \subseteq \binom{\mathcal{V}}{l}$ denotes hyperedges connecting exactly $l$ vertices; $\vw: \mathcal{E} \rightarrow \mathbb{R}$ assigns utilities to hyperedges representing expected returns from coordination; and $\mathcal{K} = \{l_{\min}, \ldots, l_{\max}\}$ specifies the range of possible hyperedge cardinalities. For each hyperedge $e \in \mathcal{E}_l$, its cardinality $|e| = l$ represents the number of coordinating agents. The open formulation allows the active subset of hyperedges to vary dynamically as team composition changes during episodes.
\end{definition}


Although the OH-Game model offers a comprehensive framework for modeling multi-scale agent interactions, directly extracting data relevant to cooperative capabilities within the game remains a challenge. To address this, we further introduce the preference hypergraph $\mathcal{OPG}$.

\begin{definition}[Open Preference Hypergraph]
An open preference hypergraph, denoted as $\mathcal{OPG}$, is an unweighted directed hypergraph derived from an OH-Game $\mathcal{G} = (\mathcal{V}, \mathcal{E}, \vw, \mathcal{L})$ and represented as $(\mathcal{V}, \widetilde{\mathcal{E}})$. For each node $i \in \mathcal{V}$ and each cardinality $l \in \mathcal{L}$, there exists an outgoing hyperedge $\widetilde{e}_i^l \in \widetilde{\mathcal{E}}$ such that $\vw(\widetilde{e}_i^l) = \max_{e \in \mathcal{E}_{i,l}} \vw(e)$, where $\mathcal{E}_{i,l} = \{e \in \mathcal{E}_l : i \in e\}$ is the set of $l$-cardinality hyperedges that connect to node $i$. This captures agent $i$'s preferred coalition partners for each possible team size.
\end{definition}

In the hyper-preference graph, a node that serves as the endpoint of multiple hyperedges across different cardinalities typically indicates higher cooperative ability and adaptability to varying team compositions. We introduce the concept of \textbf{hyper-preference centrality}, denoted by $\eta$, to quantify the cooperative ability of each node. For any node $i \in \mathcal{V}$, the hyper-preference centrality $\eta_i$ is defined as
\begin{equation}
    \eta_i = \frac{1}{(|\mathcal{V}|-1) \cdot |\mathcal{K}|} \sum_{l \in \mathcal{K}} d_l(i)
\end{equation}
where $d_l(i)$ quantifies the importance of node $i$ within the $l$-cardinality preference network, measuring how many agents prefer to coordinate with $i$ in teams of size $l$. The summation over all cardinalities $l \in \mathcal{L}$ captures an agent's overall cooperative capacity across different team scales.

Intuitively, each hyperedge in $\widetilde{\mathcal{E}}$ signifies a preference relationship, with the source node favoring the formation of a coalition of specific size with the end nodes. This preference arises because the source node achieves the highest outcomes when co-playing with the end nodes in teams of that size. Fig.~\ref{fig:pref_H} provides a schematic illustration of the hypergraph representation of OH-Game (left) and its corresponding preference hypergraph (right). In this example, nodes $v_2$ and $v_3$ exhibit the highest cooperative capacity, indicating their desirability as teammates across multiple team configurations.

Furthermore, a \textbf{best-preferred agent} is defined as one achieving hyper-preference centrality $\eta = 1$, indicating that it is the most preferred coordination partner across all team scales within the population. Intuitively, for any agent $\pi_i$, its best-preferred partner $\pi_j$ ($\pi_i \neq \pi_j$) is the one with whom $\pi_i$ achieves the highest expected return across varying team compositions, environments, and constraint satisfaction scenarios. 

\subsection{Hypergraphic Open-ended Learning Algorithm}
\label{sec:HOLA}

We then incorporate the hypergraphic-form game into open-ended learning framework and propose \algo, which continuously adjusts training objectives to enhance coordination capabilities among agents. As shown in Fig.~\ref{fig:model}, \algo consists of two main phases: the pre-training phase and the open-ended learning phase.

\subsubsection{Pre-training phase} To improve the diversity of policies in the hypergraph, we first pre-train a population of drone agents and then construct the initial OH-Game $\gG_0$. 
Building on the principles of maximum entropy reinforcement learning and maximum entropy population-based training (MEP)~\cite{MEP}, we incorporate an additional maximum entropy objective into the training function to cultivate a diverse population of strategies.
This encourages the development of a population of agents that can cooperate effectively while employing mutually distinct strategies.
The objective function is defined as follows:
\begin{equation}
J(\bar{\pi}) = \sum_t \mathbb{E}_{\left(s_t, a_t\right) \sim \bar{\pi}}\left[R\left(s_t, a_t\right) + \alpha \mathcal{H}\left(\bar{\pi}\left(\cdot \mid s_t\right)\right)\right].
\end{equation}
The objective optimizes the policy $\bar{\pi}$ by maximizing the expected reward $R(s_t, a_t)$ and encouraging exploration through the entropy term $\mathcal{H}(\bar{\pi}(\cdot \mid s_t))$, with a trade-off controlled by the weight $\alpha\in [0,1]$.
The entropy term is defined as follows:
\begin{equation}\mathcal{H}\left(\bar{\pi}\left(\cdot \mid s_t\right)\right)=-\sum_{a \in \mathcal{A}} \bar{\pi}^{(i)}\left(a_t \mid s_t\right) \log \bar{\pi}^{(i)}\left(a_t \mid s_t\right).
\end{equation}

\subsubsection{Open-ended learning phase} 
After pretraining the initial population, \algo begins continuous training by iteratively developing best-preferred agents within the evolving Open Hypergraphic-Form Game (O-HyFoG). The \algo framework consists of two main modules: the {Grapher module} constructs the evolving O-HyFoG, and the {Oracle module} trains improved agents through targeted partner sampling via its internal {$\phi$ Solver submodule}.

\paragraph{Grapher module}
The Grapher module constructs the latest O-HyFoG $\mathcal{G}_j = (\mathcal{V}_j, \mathcal{E}_j, \vw_j, \mathcal{L})$ by integrating newly generated learner agents $\Pi_j^{\text{new}} = \{\pi_j^1, \ldots, \pi_j^{n_j}\}$ with the existing population $\mathcal{V}_{j-1}$, where the learner count $n_j \sim \mathcal{P}_{\text{team}}(\{1, \ldots, n_{\max}\})$ varies across episodes to promote adaptability to dynamic team configurations. For each cardinality $l \in \mathcal{L}$ with $l \geq n_j$, the module samples $(l - n_j)$ partner agents from $\mathcal{V}_{j-1}$ to form hyperedges with $\Pi_j^{\text{new}}$. For each hyperedge $e \in \mathcal{E}_{j,l}$ connecting $l$ agents, the module evaluates coordination performance across multiple environments $\varepsilon \sim \mathcal{P}_{\text{env}}(\mathcal{E})$ and computes the average return:
\begin{equation}
\resizebox{\linewidth}{!}{
$
    \vw_j(e) = \mathbb{E}_{\varepsilon \sim \mathcal{P}_{\text{env}}}\left[\mathbb{E}_{n_j \sim \mathcal{P}_{\text{team}}}\left[\mathbb{E}_{\tau \sim \{\pi^i : i \in e\}}\left[\sum_{t=0}^T \gamma^t R(s_t, \mathbf{a}_t, \varepsilon, \phi)\right]\right]\right],
$
}
    \label{eq:hyperedge_weight}
\end{equation}
    
where $\{\pi^i : i \in e\}$ denotes the joint policy of agents in hyperedge $e$, and $R(s_t, \mathbf{a}_t, \varepsilon, \phi) = R^{\text{task}}(s_t, \mathbf{a}_t, \varepsilon) + R^{\text{constraint}}(s_t, \mathbf{a}_t, \phi)$ captures both task performance and physical constraint satisfaction. This formulation ensures hyperedge weights reflect coalition effectiveness across diverse environments and variable learner configurations.

\paragraph{Oracle module}
The Oracle module is the core of \algo, training new agents to become 
best-preferred partners across the population. Given learner policy 
$\pi_j$ and the current O-HyFoG $\mathcal{G}_j$ from the Grapher module, 
the Oracle seeks a new policy $\pi_{j+1}$ by optimizing:
\begin{equation}
    \pi_{j+1} = \textsc{BestResponse}(\pi_j, \mathcal{G}_j, \mathcal{J}_j), 
    \quad \text{such that } \eta(\pi_{j+1}) = 1,
    \label{eq:oracle}
\end{equation}
where the objective function $\mathcal{J}_j$ is defined as:
\begin{equation}
    \mathcal{J}_j = \mathbb{E}_{\substack{\pi_{-j} \sim \phi_j(\mathcal{V}_j) \\ 
    e \sim \mathcal{P}_{\text{env}} \\ \mathcal{C}_t \sim \mathcal{P}_{\text{team}}}} 
    \left[\mathbb{E}_{\tau \sim \{\pi_j, \pi_{-j}, \pi^{\text{ev}}\}}
    \left[\sum_{t=0}^T \gamma^t R(s_t, a_t, e, \phi)\right]\right],
    \label{eq:objective_open}
\end{equation}
where $\pi^{\text{ev}}$ denotes the evader policy. The three expectation 
terms capture the core dimensions of open adaptive teaming: 
$\pi_{-j} \sim \phi_j$ targets coordination with low-capability partners, 
$e \sim \mathcal{P}_{\text{env}}$ ensures generalization across diverse 
environments, and $\mathcal{C}_t \sim \mathcal{P}_{\text{team}}$ models 
within-episode team membership changes.

This objective integrates three key aspects of open adaptive teaming: partner sampling $\pi_{-j} \sim \phi_j$ targets coordination with low-capability agents, environment distribution $e \sim \mathcal{P}_{\text{env}}$ ensures generalization across diverse scenarios, and dynamic team composition $\mathcal{C}_t \sim \mathcal{P}_{\text{team}}$ captures within-episode team membership changes. The trajectory $\tau = \{(s_0, a_0, \mathcal{C}_0), (s_1, a_1, \mathcal{C}_1), \ldots, (s_T, a_T, \mathcal{C}_T)\}$ includes state-action sequences and time-varying team compositions $\mathcal{C}_t$, while the evader policy $\pi^e$ represents the target's behavior in pursuit tasks.

The key to effective training lies in the partner sampling distribution $\phi_j$, which is computed by the Oracle's internal $\phi$ Solver submodule. Given the O-HyFoG $\mathcal{G}_j$, the $\phi$ Solver derives a distribution that inversely weights agents by their cooperative capacity, ensuring that agents with lower cooperative ability receive higher sampling probability. This directs training effort toward improving coordination with challenging partners. We compute $\phi_j$ using the inverse of the generalized Myerson value adapted for hypergraphs with varying cardinalities.

For any agent $i \in \mathcal{V}_j$, the inverse Myerson value $\phi_j^{-1}(i)$ is computed as:
\begin{equation}
    \phi_j^{-1}(i) = \sum_{l \in \mathcal{L}} \frac{1}{|\Pi(\mathcal{V}_j)|} \sum_{\sigma \in \Pi(\mathcal{V}_j)} [v_l(\mathcal{P}_i^\sigma \cup \{i\}) - v_l(\mathcal{P}_i^\sigma)],
    \label{eq:myerson_multi}
\end{equation}
where $\Pi(\mathcal{V}_j)$ is the set of all permutations of $\mathcal{V}_j$, $\mathcal{P}_i^\sigma$ denotes the set of agents preceding $i$ in permutation $\sigma$, and $v_l(S)$ is the coalition value function for coalitions of size $l$:
\begin{equation}
    v_l(S) = \begin{cases}
        \sum_{e \in \mathcal{E}_{j,l}, e \subseteq S} \vw_j(e), & \text{if } |S| \geq l, \\
        0, & \text{if } |S| < l.
    \end{cases}
    \label{eq:coalition_value}
\end{equation}

The summation over $l \in \mathcal{L}$ in Eq.~\ref{eq:myerson_multi} aggregates contributions across all team scales, capturing an agent's overall cooperative value in open settings.

\begin{proposition}[Myerson Value for O-HyFoG]
For a connected O-HyFoG $\mathcal{G} = (\mathcal{V}, \mathcal{E}, \vw, \mathcal{L})$, the inverse Myerson value can be equivalently computed as:
\begin{equation}
    \phi^{-1}(i) = \sum_{l \in \mathcal{L}} SV_i^l(\mathcal{V}, v_l),
    \label{eq:myerson_shapley}
\end{equation}
where $SV_i^l(\mathcal{V}, v_l)$ is the Shapley value of agent $i$ in the $l$-cardinality coalition game.
\end{proposition}

For each cardinality $l \in \mathcal{L}$, we show that the contribution to $\phi^{-1}(i)$ from $l$-coalitions equals the Shapley value $SV_i^l(\mathcal{V}, v_l)$. Since O-HyFoG is connected, any subset $S \subseteq \mathcal{V}$ induces a connected subhypergraph. For a given permutation $\sigma \in \Pi(\mathcal{V})$, define the marginal contribution of agent $i$ in $l$-coalitions as:
\begin{equation}
    \Delta_l(i, \sigma) = v_l(\mathcal{P}_i^\sigma \cup \{i\}) - v_l(\mathcal{P}_i^\sigma).
\end{equation}

By definition of $v_l$ in Eq.~\ref{eq:coalition_value}, we have:
\begin{align}
    \Delta_l(i, \sigma) &= \sum_{e \in \mathcal{E}_l, e \subseteq \mathcal{P}_i^\sigma \cup \{i\}} \vw(e) - \sum_{e \in \mathcal{E}_l, e \subseteq \mathcal{P}_i^\sigma} \vw(e) \\
    &= \sum_{e \in \mathcal{E}_l : i \in e, e \setminus \{i\} \subseteq \mathcal{P}_i^\sigma} \vw(e),
\end{align}
which counts only those $l$-cardinality hyperedges that include $i$ and have all other members in $\mathcal{P}_i^\sigma$. Averaging over all permutations yields:
\begin{equation}
    \frac{1}{|\Pi(\mathcal{V})|} \sum_{\sigma \in \Pi(\mathcal{V})} \Delta_l(i, \sigma) = SV_i^l(\mathcal{V}, v_l),
\end{equation}
by the definition of Shapley value. Summing over all cardinalities $l \in \mathcal{L}$ establishes Eq.~\ref{eq:myerson_shapley}.

The sampling distribution $\phi_j$ is then computed as:
\begin{equation}
    \phi_j(i) = \frac{\phi_j^{-1}(i)^{-1}}{\sum_{k \in \mathcal{V}_j} \phi_j^{-1}(k)^{-1}},
    \label{eq:sampling_dist}
\end{equation}
ensuring higher probability for agents with lower cooperative capacity.

The constraint $\eta(\pi_{j+1}) = 1$ in Eq.~\ref{eq:oracle} ensures that $\pi_{j+1}$ becomes the best-preferred agent, achieving the highest hyper-preference centrality across all team scales. However, this strict condition may be infeasible. Thus, we relax it to require that $\pi_{j+1}$'s centrality rank within the top $m$:
\begin{equation}
\begin{aligned}
    &\pi_{j+1} = \textsc{ApproxBestResponse}(\pi_j, \mathcal{G}_j, \mathcal{J}_j), \\
    &\quad \text{such that } \text{rank}(\eta(\pi_{j+1})) \leq m,
\end{aligned}
    \label{eq:approx_oracle}
\end{equation}
where $\text{rank}(\cdot)$ returns the ranking of centrality scores. The resulting policy $\pi_{j+1}$ is termed an {approximate best-preferred agent}.

The complete procedure is presented in Algorithm~\ref{alg:hola-drone}, and Fig.~\ref{fig:model} illustrates the workflow for a single generation.
\begin{algorithm}[t]
\caption{Open-Ended Learning Phase of \algo}
\label{alg:hola-drone}
\begin{algorithmic}[1]
    \State \textbf{Input:} Pre-trained population $\mathcal{G}_0$, cardinality set $\mathcal{L}$, robot constraints $\Phi$
    \State \textbf{Input:} Environment distribution $\mathcal{P}_{\text{env}}$, team size distribution $\mathcal{P}_{\text{team}}$, top-$m$ threshold

    \For{each generation $j \in \{1, 2, \ldots\}$}
        \State \Comment{\textbf{Grapher Module}}
        \State Sample learner count: $n_j \sim \mathcal{P}_{\text{team}}(\{1, \ldots, n_{\max}\})$
        \State $\mathcal{V}_j \gets \mathcal{V}_{j-1} \cup \Pi_j^{\text{new}}$
        \For{each cardinality $l \in \mathcal{L}$ with $l \geq n_j$}
            \For{each subset $S \subseteq \mathcal{V}_{j-1}$ with $|S| = l - n_j$}
                \State Form hyperedge $e \gets S \cup \Pi_j^{\text{new}}$
                \State Sample environments $\{\varepsilon_1, \ldots, \varepsilon_K\} \sim \mathcal{P}_{\text{env}}$
                \State Evaluate: 
                \State \resizebox{0.8\linewidth}{!}{$\vw_j(e) \gets \frac{1}{K}\sum_{k=1}^K\mathbb{E}_{\tau}\left[\sum_{t=0}^T \gamma^t R(s_t,\mathbf{a}_t,\varepsilon_k,\phi)\right]$}
                \State $\mathcal{E}_{j,l} \gets \mathcal{E}_{j,l} \cup \{e\}$
            \EndFor
        \EndFor
        \State $\mathcal{G}_j \gets (\mathcal{V}_j, \bigcup_{l \in \mathcal{L}} \mathcal{E}_{j,l}, \vw_j, \mathcal{L})$
        
        \State \Comment{\textbf{Oracle Module}}
        \For{each agent $i \in \mathcal{V}_j$}
            \State Compute $\phi_j^{-1}(i)$ via Eq.~\ref{eq:myerson_multi}
        \EndFor
        \State $\phi_j(i) \gets \phi_j^{-1}(i)^{-1} / \sum_{k} \phi_j^{-1}(k)^{-1}$ for all $i \in \mathcal{V}_j$
        
        \Repeat
            \State Sample learner count: $n_{j+1} \sim \mathcal{P}_{\text{team}}$
            \State Sample partners: $\pi_{-j} \sim \phi_j$ with $|\pi_{-j}| = l - n_{j+1}$ for sampled $l \in \mathcal{L}$
            \State Sample environment: $\varepsilon \sim \mathcal{P}_{\text{env}}$
            \State Update learners: 
            \State $\Pi_{j+1}^{\text{new}} \gets \textsc{Update}(\Pi_j^{\text{new}}, \pi_{-j}, \varepsilon, \phi)$ via Eq.~\ref{eq:objective_open}
        \Until{$\text{rank}(\eta(\Pi_{j+1}^{\text{new}})) \leq m$}
    \EndFor
    \State \textbf{Return:} Trained population $\mathcal{G}_j$
\end{algorithmic}
\end{algorithm}
\section{Experimental Setting}
\label{sec:exp_setting}

We validate our approach on cooperative multi-robot pursuit tasks using two robotic platforms: multi-drone and multi-quadruped systems. The experiments are designed to evaluate the three core dimensions of open adaptive teaming: coordination with unseen teammates, generalization to unseen environments, and scalability to variable team sizes. We first describe the task formulation and platform specifications, then detail the evaluation protocols including both open team and fixed team settings. Finally, we present the evader policy, and unseen teammate pools.

\subsection{Task Formulation}
We adopt multi-robot cooperative pursuit as the primary testbed for evaluating open adaptive teaming. In this task, a team of pursuers must coordinate to capture multiple evaders within a bounded arena containing obstacles. The pursuit scenario is particularly well-suited for assessing open adaptive teaming capabilities because it inherently demands real-time coordination under dynamic conditions: pursuers are typically slower than evaders, making individual capture infeasible and necessitating strategic cooperation that must continuously adapt as team composition and environmental configurations change.

At the start of each episode, pursuers and evaders are randomly spawned in designated regions of the arena. An evader is considered captured when any pursuer moves within the capture distance $d_c$. Each agent operates under limited perception with range $d_p$ and maintains a safe radius $d_s$ for collision avoidance. The episode terminates when all evaders are captured or the maximum time horizon $t_{max}$ is reached.

\subsection{Robotic Platforms and Real-World Deployment}
\label{subsec:platforms}

\begin{figure}
    \centering
    \includegraphics[width=1\linewidth]{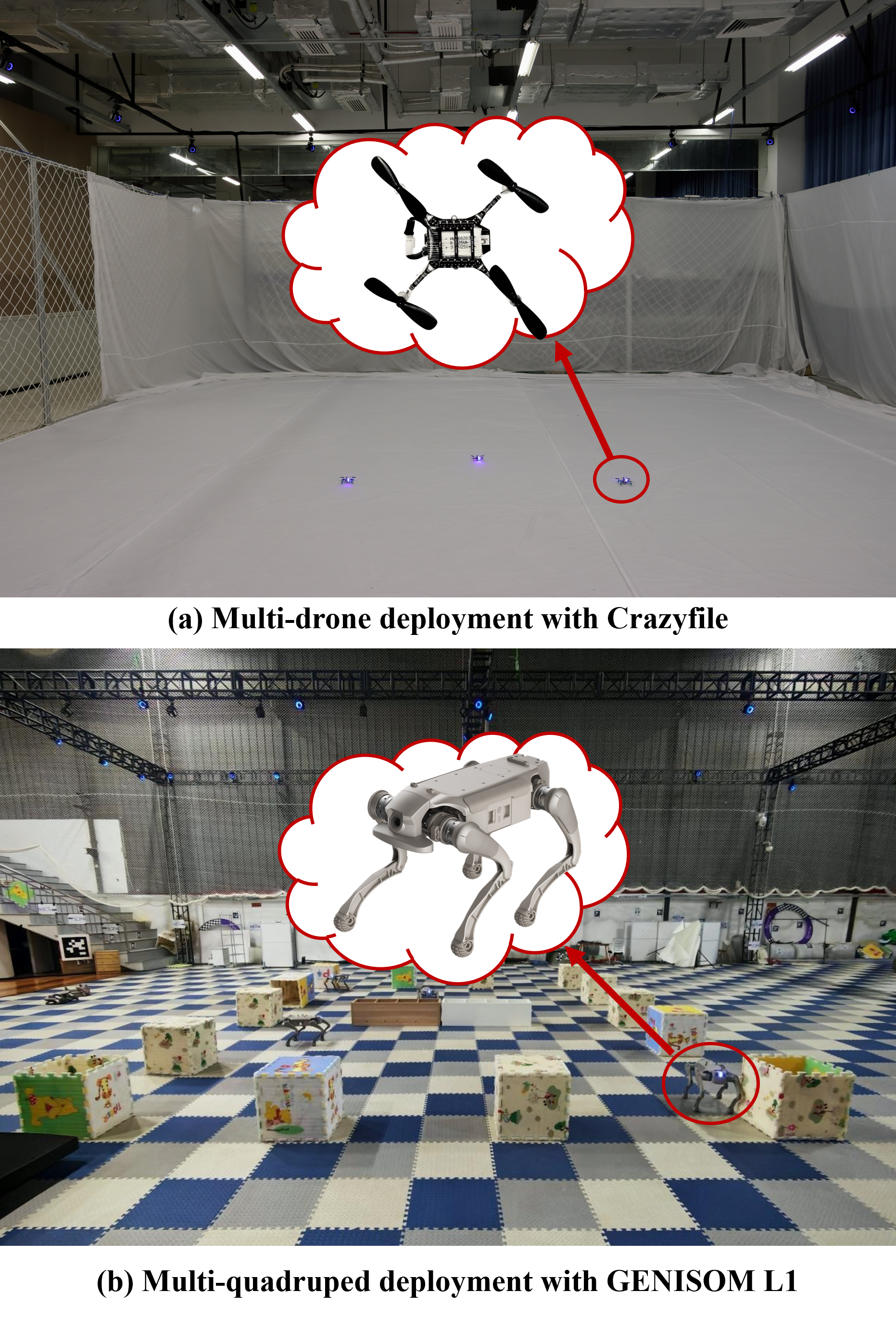}
    \caption{Real-world deployment of multi-robot open adaptive teaming on two robotic platforms.}
    \label{fig:deployment}
\end{figure}

To validate the generality of our approach across different embodiments, we deploy the cooperative pursuit task on two distinct robotic platforms: a multi-drone system and a multi-quadruped system. While both platforms implement the same task formulation, they differ fundamentally in physical dynamics, motion constraints, and operational scales. The multi-drone system operates in aerial space with high maneuverability but limited payload, whereas the multi-quadruped system navigates the ground plane with different inertial properties and contact dynamics. This diversity in platform characteristics enables robust evaluation of whether the learned coordination strategies generalize across embodiment types. Both platforms share a common system architecture: a motion capture system provides global pose estimation, a centralized pursuit planner executes the learned policy, and wireless communication modules relay control commands to individual robots.

\subsubsection{Multi-Drone Platform}
As shown in Fig.~\ref{fig:deployment}, the drone experiments are conducted using a Crazyflie 2.1 multi-quadrotor system within a rectangular flight arena measuring $w_b = 3.6$\,m in width and $h_b = 5.0$\,m in length. Each episode has a time horizon of $t_{\max} = 100$ seconds, and the system runs at 10\,fps. The training team configuration consists of three pursuers ($p_1, p_2, p_3$) and two evaders ($e_1, e_2$). At episode initialization, agents are randomly spawned in designated areas of $w_s = 3.2$\,m $\times$ $h_s = 0.6$\,m, with evaders in the upper region and pursuers in the lower region. The arena contains 5 obstacles, each measuring $w_o = 0.65$\,m $\times$ $h_o = 0.1$\,m. The capture distance is $d_c = 0.2$\,m, the perception range is $d_p = 2.0$\,m, and each drone maintains a safe radius $d_s = 0.1$\,m for collision avoidance. Pursuers move at $v_P = 0.3$\,m/s while evaders move at $v_E = 0.6$\,m/s, creating a 2:1 speed ratio that necessitates coordinated encirclement.

For real-world deployment, drone positions are tracked by a network of 12 FZMotion cameras operating at 120\,Hz. The position data, represented as point clouds, is collected by a motion capture server via USB and transmitted to the pursuit planner over Ethernet. The planner runs on a Lenovo ThinkPad T590 equipped with a Jetson Orin Nano module, which performs policy inference and sends velocity commands to each drone through a Crazyradio PA wireless dongle. On each Crazyflie, an onboard Mellinger controller tracks the received velocity commands. Control commands are issued at approximately 10\,Hz to ensure stable flight. The communication and control pipeline is built upon the CrazySwarm framework.

\subsubsection{Multi-Quadruped Platform}
The quadruped experiments employ the L1 robot developed by GEMBODY AI, operating in a ground-based rectangular arena measuring $6.0$\,m in width and $8.0$\,m in length. The task horizon and frame rate remain identical to the drone platform at $t_{\max} = 100$ seconds and 10\,fps, with the same team composition of 3 pursuers and 2 evaders. Spawn areas maintain proportional dimensions relative to the larger arena. The capture distance is increased to $d_c = 0.65$\,m to account for the physical dimensions of the quadruped robots, while the perception range remains $d_p = 2.0$\,m. The safe radius is set to $d_s = 0.4$\,m, reflecting the larger footprint of ground robots. Movement velocities are configured identically to the drone platform with $v_P = 0.3$\,m/s and $v_E = 0.6$\,m/s, preserving the same speed differential across platforms.

The real-world arena
, shown in Fig.~\ref{fig:deployment}, is instrumented with a high-precision spatial positioning system consisting of 50 motion capture cameras that provide real-time 6-DoF pose estimation for all robots. Pose data is acquired by a motion capture server via USB and forwarded to the pursuit planner through Ethernet. After data parsing and fusion, the planner distributes position control commands to each robot via wireless communication modules at a rate of 10\,Hz. At the communication level, a distributed architecture based on ROS2 and the Chrony time synchronization protocol achieves end-to-end latency of $\leq$\,10\,ms. At the motion control level, each L1 robot runs a locomotion policy trained via deep reinforcement learning, which provides robust dynamic walking capability over the foam-mat terrain used in the arena.

\subsection{Evaluation Protocols}
We evaluate our approach under two experimental protocols that systematically probe different aspects of open adaptive teaming.

\subsubsection{Open Team Protocol} 
This protocol implements the complete open adaptive teaming framework across all three dimensions. First, to test coordination with unseen teammates, we construct a diverse partner pool $\mathcal{U} = \{u_1, \dots, u_K\}$ where each policy is trained with a distinct algorithm exhibiting different cooperative capabilities. At each episode, teammates are sampled from $\mathcal{U}$, and team composition may change within episodes as partners join or leave. The learning agent has no prior knowledge of teammate policies and must coordinate without pre-established protocols. Second, to test generalization to unseen environments, the number, size, and placement of obstacles are randomized to produce spatial configurations not encountered during training. Third, to test scalability across variable team sizes, the number of active pursuers varies both across and within episodes, ranging from small teams requiring tight coordination to larger teams demanding scalable cooperation mechanisms.

\subsubsection{Fixed Team Protocol} 
As a controlled ablation study, we also evaluate performance under a fixed team setting on the multi-drone platform. In this protocol, team size is held constant at 3 pursuers and 2 evaders throughout each episode, while the unseen teammate and unseen environment conditions remain active. By comparing fixed and open team results on the same platform, we isolate the impact of variable team size adaptation and quantify its contribution to overall performance.

\subsection{Evader Policy} 
The evaders are controlled by the escape policy proposed by \cite{Janosov2017Group} and \cite{ZhangDACOOP2023}. This policy defines multiple repulsive forces exerted by pursuers and obstacles on evaders. Additionally, wall-following rules are incorporated to help evaders maneuver along obstacle surfaces when positioned between pursuers and obstacles.

\begin{table}[t]
\centering
\caption{
One-evader capture success rate (SR) and average episode length (AEL) performance of agents in the unseen heterogeneous teammate pool. 
The numbers (1) and (2) following D3QN-G represent models trained with different random seeds. 
All results are averaged over 50 validation episodes.
}
\resizebox{\linewidth}{!}{
\begin{tabular}{ccccc}
    \toprule
    \textbf{Metrics} & \textbf{Greedy} & \textbf{VICSEK} & \textbf{D3QN-G (1)} & \textbf{D3QN-G (2)}  \\
    \midrule
    \textbf{SR} & 62.0\% & 98.0\% & 80.0\% & 78.0\%  \\
    \textbf{AEL} & 561.78 & 295.88 & 435.78 & 510.34  \\
    \bottomrule
\end{tabular}
}
\label{tab:unseen_pool}
\end{table}

\subsection{Unseen Teammate Pools}
\label{sec:exp_unseen}

To evaluate cooperative capabilities with previously unencountered partners, we construct a heterogeneous pool consisting of four distinct models:
{Greedy Agent.} This agent pursues targets independently, continually adjusting movement to align with evader positions. State information includes own position and orientation, distances and angles to teammates and evaders, and proximity to obstacles. The agent applies evasive maneuvers when obstacles or pursuers are detected within its avoidance range.
{VICSEK Agent.} Inspired by group chasing tactics~\cite{Janosov2017Group}, this strategy continuously computes velocity vectors directed toward evaders based on current environmental state. When obstacles or other pursuers are detected nearby, the agent applies repulsive forces with varying magnitudes. Although the computed velocity vector includes both magnitude and orientation, only the orientation component is utilized in the current implementation.
{D3QN-G Agent.} This ensemble algorithm combines Double Deep Q-Network (D3QN)~\cite{wang2016dueling} with the Greedy strategy. Initially, D3QN-G employs the D3QN method to pursue the first evader. Upon successful capture, it switches to the Greedy strategy for the second evader. The action space consists of 24 artificial potential field with attention (APF-A) parameter pairs ($\lambda$, $\eta$), formed by the Cartesian product of 8 $\lambda$ candidates and 3 $\eta$ candidates, following~\cite{ZhangDACOOP2023}. The parameter $\eta$ calculates repulsive force while $\lambda$ determines inter-robot force. The state dimension is 9, including position, orientation, teammate and evader information, obstacle proximity, and an activity bit indicating whether the agent remains active after teammate captures.

\begin{figure*}[ht!]
    \centering
    \vspace{-0.5cm}\includegraphics[width=\linewidth]{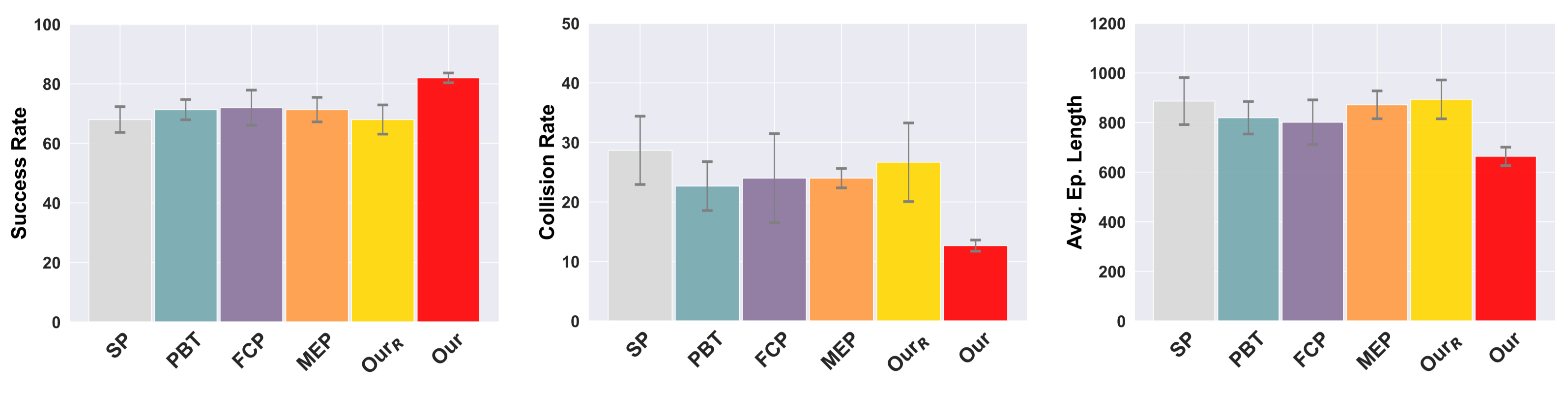}
\caption{
    Comparison of task success rate (first column, higher is better), 
    collision rate (second column, lower is better), and mean episode 
    length (third column, lower is better) among four baseline methods, 
    one ablation method \algoR\ (marked as Our$_R$), and our proposed 
    \algo in the 3-pursuer-2-evader scenario when playing with random 
    heterogeneous teammates (where policies are derived from different 
    algorithms). The means and standard deviations, indicated by the 
    error bars, are calculated over three different random seeds, with 
    each seed undergoing 50 repeated runs.
    }
    \vspace{-0.4cm}
    \label{fig:main_res}
\end{figure*}

As shown in Table~\ref{tab:unseen_pool}, these agents span a capability spectrum from low-level Greedy (62\% SR), through medium-level D3QN-G variants (approximately 80\% SR), to expert-level VICSEK (98\% SR). At each evaluation episode, we randomly sample two teammates from the pool to test zero-shot coordination ability across varying skill levels.

\section{Experimental Results}
\label{sec:exp_res}

We evaluate \algo through a progressive experimental protocol that validates both fundamental coordination capabilities and comprehensive open adaptive teaming performance. The fixed team experiments establish baseline coordination abilities under controlled conditions with constant team composition within episodes. Building on this foundation, the open team experiments evaluate the core challenge of adaptive teaming where team composition changes dynamically during task execution.

\subsection{Fixed Team Coordination}

Fixed team experiments assess zero-shot coordination capabilities where team composition remains constant within each episode, allowing systematic evaluation of adaptation to unseen teammates and novel environments.

\subsubsection{Coordination with Unseen Teammates and Environments}

Fig.~\ref{fig:main_res} compares \algo performance against six baseline methods: MAPPO~\cite{mappo}, DACOOP-A~\cite{ZhangDACOOP2023}, self-play~\cite{schulman2017equivalence,carroll2019utility}, PBT~\cite{jaderberg2017population,carroll2019utility}, FCP~\cite{FCP}, and MEP~\cite{MEP}. We construct a heterogeneous teammate pool containing four drone agents with varying cooperative abilities from medium to expert levels. For each evaluation episode, two teammates are randomly sampled from this pool, repeated 50 times per seed to ensure statistical robustness.

Results reveal fundamental limitations of traditional CTDE-based approaches. MAPPO and DACOOP-A achieve no more than 40\% success rate, 
demonstrating that methods trained under fixed team configurations 
fail to generalise to unseen partners, regardless of their 
underlying coordination architecture.
In contrast, \algo reaches 82\% success rate with a 13.67\% collision rate, substantially outperforming FCP (72\% success, second-best) and PBT (22.67\% collision, highest among baselines). The efficiency advantage is equally pronounced, with \algo a required mean episode length of 663.72 timesteps compared to 801.19 for the closest competitor FCP.

\begin{table}[t]
    \centering
    \caption{Performance comparison for generalization to randomly sampled novel environments. Results are averaged over 50 repeated experiments with three different random seeds, where each experiment evaluates performance in a newly sampled unseen environment from the pool.}
    \resizebox{\linewidth}{!}{
    \begin{tabular}{cccccccc}
\toprule 
 \multirow{2}{*}{\textbf{Metrics}} & 
    \multicolumn{7}{c}{\textbf{Methods}}  \\
    \cmidrule(lr){2-8} 
    & \textbf{MAPPO}
     & \textbf{DACOOP-A} & \textbf{SP} & \textbf{PBT} & \textbf{FCP} & \textbf{MEP} & \textbf{\algo} \\
\midrule
\textbf{SUC} & 35.33  & 25.33 & 40.00 & 42.67  & 40.00 & 36.00 & \textbf{44.00} \\
\textbf{AST} & 1388.11 & 1553.53 & 1316.05& \textbf{1290.91} & 1318.02 & 1392.43 & 1294.65\\
\textbf{COL} & 61.33 & 70.00 & 58.00  & \textbf{54.67} & 56.67 & 61.33  & \textbf{54.67} \\
\bottomrule
\end{tabular}
}
    \label{tab:exp_gen}
\end{table}

Table~\ref{tab:exp_gen} reports performance in randomly generated unseen environments to assess simultaneous adaptation across both teammate and environment dimensions. During training, agents operate in fixed layouts with five obstacles. The evaluation pool contains diverse configurations with three or four randomly placed rectangular obstacles, with a new environment sampled at the start of each experiment.

Averaged over 50 experiments across three random seeds, \algo achieves the highest success rate (44.00\%) and matches the lowest collision rate (54.67\%). For average success timesteps, \algo ranks second at 1294.65, trailing PBT by fewer than four steps while maintaining superior success and collision metrics. These results confirm that \algo effectively handles coordination uncertainty across both partner and environment dimensions in fixed team settings.

\subsubsection{Ablation Study: Effectiveness of $\phi$ Solver Module}

To isolate the contribution of the $\phi$ solver module, we evaluate \algoR, a variant that removes this component and instead uses inverse mean reward as the cooperative relationship distribution $\phi$. Fig.~\ref{fig:main_res} (gold bars) shows consistent performance degradation across all metrics when the $\phi$ solver is removed. Compared to full \algo (red bars), \algoR exhibits lower success rates, higher collision rates, and longer episode lengths in heterogeneous teammate configurations. This degradation validates that explicit modeling of team-level cooperative dependencies through the hypergraphic formulation is essential for effective coordination with unseen partners.

\begin{figure*}[ht]
    \centering
    \subfloat[Multi-drone open adaptive teaming]{%
        \includegraphics[width=\linewidth]{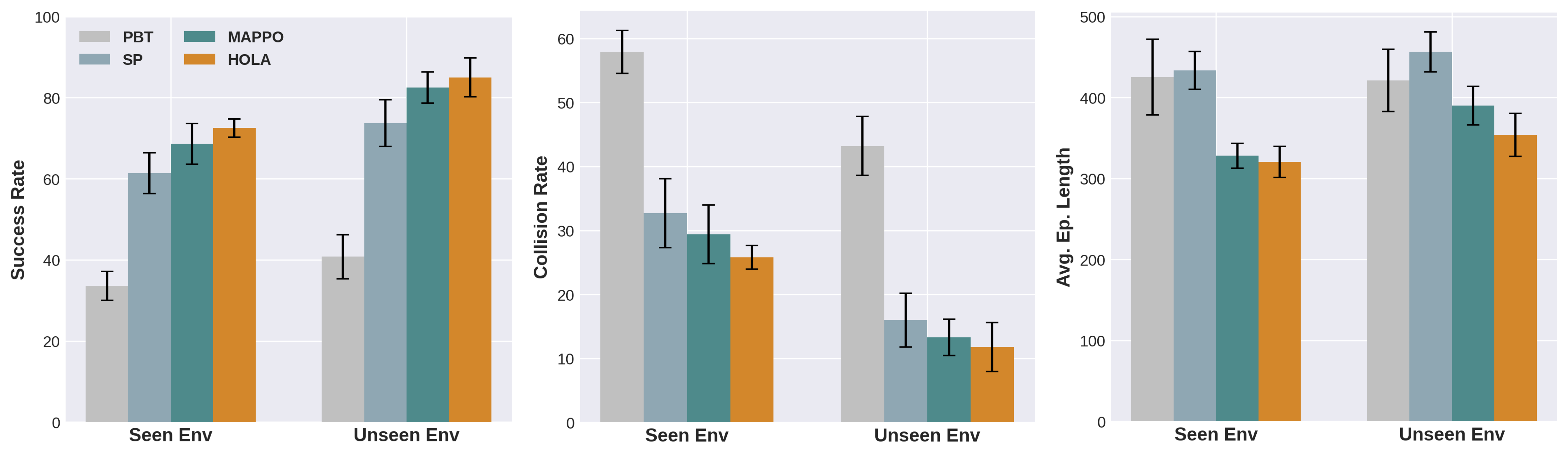}%
        \label{fig:open_res_drone}%
    }\\
    \subfloat[Multi-quadruped open adaptive teaming]{%
        \includegraphics[width=\linewidth]{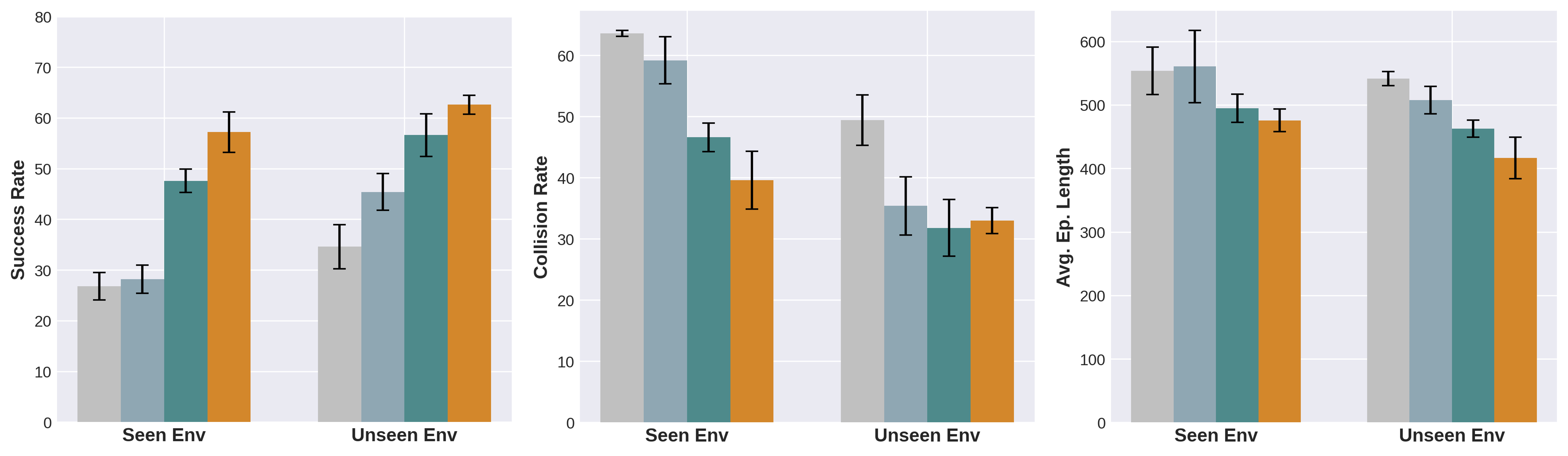}%
        \label{fig:open_res_dog}%
    }
    \caption{Open team coordination on multi-drone (top row) and multi-quadruped 
    (bottom row) platforms. Metrics: capture rate ($\uparrow$), collision rate 
    ($\downarrow$), average episode length ($\downarrow$), evaluated on seen 
    (one-obstacle) and unseen (three-obstacle) environments. \algo leads on 
    capture rate and episode efficiency across all conditions, and generalizes 
    favorably to unseen layouts where denser obstacle topology creates strategic 
    cornering opportunities.}
    \label{fig:open_res}
\end{figure*}

\subsection{Open Team Coordination}

The open team protocol evaluates the fundamental capability of adaptive teaming by introducing dynamic team composition changes within episodes, not merely across episodes. This represents a critical departure from fixed team settings, requiring agents to continuously adapt as teammates join or leave during task execution while simultaneously handling novel environments.

\subsubsection{Multi-Drone Pursuit}

The top row of Fig.~\ref{fig:open_res} presents multi-drone pursuit results under the open adaptive teaming setting. We evaluate performance in both seen environments (one obstacle layout) and unseen environments (three obstacle layouts) to assess generalization capabilities.

\algo consistently achieves the highest capture rates across all conditions, reaching approximately 73\% in seen environments and further improving to around 85\% in unseen environments. {A striking observation is that most methods exhibit higher capture rates in unseen environments than in seen ones, which appears counterintuitive at first glance}. We attribute this to the richer obstacle topology in three-obstacle layouts, which creates more opportunities for strategic cornering of the evader. However, the magnitude of this improvement varies significantly across methods. \algo and MAPPO transition relatively smoothly (approximately 12 and 15 percentage point gains, respectively), while self-play shows a larger jump from 60\% to 73\%, suggesting that self-play benefits more from environmental structure 
but lacks the stable coordination needed to perform consistently 
in sparser environments.
PBT, by contrast, remains below 40\% in both conditions, confirming that population-based training alone is insufficient for open team coordination where teammates are non-stationary.

Efficiency and safety metrics reinforce these findings. \algo completes tasks in roughly 320 timesteps in both seen and unseen environments, exhibiting remarkable stability across conditions. This consistency contrasts sharply with self-play, whose completion time drops from 440 to 350 steps when moving to unseen environments, mirroring its dependence on environmental structure rather than learned coordination. Collision rates further differentiate the methods: \algo achieves approximately 14\% in unseen environments, nearly half that of self-play (24\%) and a third of PBT (44\%). The simultaneous achievement of highest capture rate and lowest collision rate demonstrates that \algo does not trade safety for aggressiveness but instead discovers coordinated pursuit strategies that are inherently more efficient.

\subsubsection{Multi-Quadruped Pursuit}

The bottom row of Fig.~\ref{fig:open_res} validates \algo's adaptability on quadruped platforms, where the coordination challenge is compounded by the complexity of legged locomotion and tighter spatial constraints. Overall performance across all methods is notably lower than on drones, reflecting the increased difficulty of this morphology.

\algo achieves 57.2\% and 62.6\% capture rates in seen and unseen environments respectively, maintaining a consistent margin of roughly 6 to 10 percentage points over the closest baseline MAPPO. The gap widens dramatically against self-play and PBT, both of which fall below 35\% in seen environments. This ordering reveals a clear hierarchy: methods with centralized training and explicit cooperative modeling (HOLA, MAPPO) substantially outperform those relying on implicit coordination (self-play, PBT), and the hypergraphic formulation in \algo provides additional gains by capturing team-level relationships beyond pairwise interactions.

An important finding emerges from the efficiency analysis. \algo completes tasks in 475.7 and 416.3 timesteps in seen and unseen conditions, maintaining approximately a 60-step advantage over MAPPO and over 80 steps faster than self-play and PBT across both settings. This consistent efficiency gap suggests that the Oracle module generates more decisive cooperative plans that reduce redundant exploration, even when team membership is changing dynamically throughout the episode.

Collision rates on quadrupeds are substantially higher than on drones across all methods, with PBT reaching 63.6\% in seen environments. \algo records 39.6\% and 33.0\% in the two conditions, generally the lowest among all methods. An interesting nuance is that MAPPO achieves a slightly lower collision rate (31.8\%) than \algo (33.0\%) in unseen environments, yet at the cost of a 6-percentage-point deficit in capture rate. This tradeoff indicates that MAPPO adopts a more conservative strategy in novel environments, whereas \algo maintains aggressive but coordinated pursuit that yields higher task success without proportionally increasing collision risk.

Taken together, the results across both robotic platforms and environmental conditions establish three key conclusions. First, the hypergraphic formulation in \algo enables robust open team coordination that generalizes across robotic morphologies, environmental layouts, and dynamic team compositions. Second, the performance advantage of \algo is not merely incremental but reflects a qualitative difference in coordination capability, particularly evident in the simultaneous optimization of capture efficiency and collision avoidance. Third, the consistency of \algo's performance across seen and unseen conditions suggests that the learned cooperative structures capture transferable relational patterns rather than environment-specific strategies, a property essential for practical deployment in unstructured real-world scenarios.

\section{Conclusion}
\label{sec:conclusion} 
This paper presents HOLA, a hypergraphic open-ended learning algorithm 
for open adaptive multi-robot teaming, which jointly addresses three 
dimensions of adaptability: coordination with unseen partners, 
generalization to novel environments, and adaptation to variable team 
sizes. Central to our approach is a hypergraphic-form game formulation 
that models team-level cooperative relationships beyond pairwise 
interactions, providing a principled basis for credit assignment and 
partner sampling in open teams.

We evaluate HOLA on cooperative multi-robot pursuit across two robotic platforms, multi-drone and multi-quadruped systems, under both fixed and open team protocols. In fixed team settings, HOLA achieves the highest success rate and lowest collision rate among all baselines, confirming its zero-shot coordination capability with unseen partners across novel environments. In open team settings, where team composition changes dynamically \textit{within} episodes rather than only across them, HOLA consistently outperforms all baselines on capture rate and episode efficiency across both seen and unseen environments. Critically, learned policies transfer directly to physical platforms without fine-tuning, with successful real-world deployments on both Crazyflie multi-drone and Zsibot L1 quadruped systems demonstrating robust coordination with unseen teammates in novel environments.

Despite these gains, several limitations remain. First, safety-aware behavior can still be improved in dense interactions, especially when agents trade collision risk for faster capture. Second, although variable team-size adaptation is validated, larger-scale teams and longer-horizon tasks remain challenging. Third, our evaluations focus on pursuit scenarios; broader multi-robot task families should be explored to assess generality further. Future work will integrate stronger safety constraints, improve scalable training and inference for larger teams, and extend open adaptive teaming to more diverse real-world cooperative tasks.

\bibliographystyle{ieeetr}
\bibliography{example}  
\end{document}